\documentclass[twoside,11pt]{article}

\usepackage{jmlr2e}
\RequirePackage[OT1]{fontenc}
\RequirePackage{amsmath}

\usepackage{amssymb}
\usepackage{float}
\usepackage{epsfig}
\usepackage{amsfonts}
\usepackage{latexsym}
\usepackage{color}
\usepackage{algorithm, algorithmic}
\usepackage[tight]{subfigure}
\usepackage{verbatim}
\usepackage{graphicx}
\usepackage{algorithm, algorithmic}
\usepackage{multirow}
\usepackage{epstopdf}
\newcommand{\R}{\mathbb{R}}
\newcommand{\Var}{\text{Var}}

\jmlrheading{1}{2000}{1-48}{4/00}{10/00}{Conroy et al.}

\ShortHeadings{FaSTGLZ}{Conroy et al.}
\firstpageno{1}

\begin{document}

\title{Fast Simultaneous Training of Generalized Linear Models (FaSTGLZ)}

\author{\name Bryan R. Conroy \email bc2468@columbia.edu
		\AND
	    \name Jennifer M. Walz \email jw2552@columbia.edu
	    	\AND
	   \name Brian Cheung \email cheung4@cooper.edu
	    	\AND
	    \name Paul Sajda \email psajda@columbia.edu\text{ }\\
	    \addr Department of Biomedical Engineering \\
	    Columbia University \\
	    New York, NY 10027 USA}
\editor{}
		
\maketitle

\begin{abstract}%   <- trailing '%' for backward compatibility of .sty file
We present an efficient algorithm for simultaneously training sparse generalized linear models across many related problems, which may arise from bootstrapping, cross-validation and nonparametric permutation testing.  Our approach leverages the redundancies across problems to obtain significant computational improvements relative to solving the problems sequentially by a conventional algorithm.  We demonstrate our fast simultaneous training of generalized linear models (\emph{FaSTGLZ}) algorithm on a number of real-world datasets, and we run otherwise computationally intensive bootstrapping and permutation test analyses that are typically necessary for obtaining statistically rigorous classification results and meaningful interpretation.  Code is freely available at http://liinc.bme.columbia.edu/fastglz.
\end{abstract}

%\linenumbers

\section{Introduction}

In machine learning, optimization algorithms are often tuned to efficiently learn a single model from data.  In reality though, a typical machine learning application involves training thousands of models on a single dataset over the course of model selection, model comparison, and statistical significance testing.  Although these optimization problems tend to be highly related to one another, it is common to solve them sequentially in a loop, or in parallel if a computing cluster is available.  This ignores the potential to exploit the present redundancies across problems to further improve computational efficiency.

This paper presents a computationally and memory efficient algorithm for simultaneously training a set of sparse regression models on a common dataset.  Throughout this paper, we refer to each individual model fit as a problem.  These problems may arise from bootstrapping, cross-validation, and permutation testing.  We show that by solving the set simultaneously as a group, we are capable of leveraging the shared structure to obtain significant computational savings.  Our algorithm, Fast Simultaneous Training of Generalized Linear Models (\emph{FaSTGLZ}), applies to a wide array of machine learning algorithms, but we focus primarily on generalized linear models (GLZ) regularized by the elastic net \citep{Tibshirani96,Zou05}.  Such models are flexible and arise in many scenarios, since GLZ's are compatible with many popular probability distributions, and readily extend to classification problems using logistic regression. Moreover, elastic net regularization allows for sparse and parsimonious solutions while avoiding the saturation problems of lasso when the number of features $p$ exceeds the number of examples $n$ \citep{Zou05}.  This is a common scenario in many real-world datasets.

Our \emph{FaSTGLZ} algorithm builds on the algorithmic framework of the alternating direction method of multipliers (ADMM) optimization procedure \citep{Eckstein92}. ADMM uses variable splitting to divide the sparse regression optimization into two simpler sub-procedures:  (a) minimizing a differentiable objective; and (b) a univariate soft-thresholding operation.  Within this framework, our approach leverages the shared structure across problems in two key ways.  First, we show that the minimization in (a) can be achieved in a low-dimensional space that is common to all problems.  Additionally, we formulate an efficient Newton solver that simultaneously minimizes (a) across problems using only one template Hessian matrix inversion. The simultaneous Newton solver has the added benefit of bundling the iterative steps into a single linear algebraic expression.  This greatly reduces overhead and memory access times.

Our algorithm is also memory efficient by incorporating the $\ell_1$-regularized screening tests of \citet{Tibshirani12} to estimate the active set of each regression problem.  We derive an expression for the amount of memory required by \emph{FaSTGLZ} per optimization problem and show that it grows linearly with the number of examples $n$, and does not grow with the number of features $p$.  In $p\gg n$ scenarios this is a substantial memory overhead reduction, and allows us to, on a standard quad-core machine with $4$G RAM, simultaneously train thousands of related sparse regression problems in a high-dimensional setting (tens of thousands of features).

The remainder of this paper is organized as follows.  In the following section, we provide background on generalized linear models and formulate the problem we seek to solve.  This is followed by a derivation of the \emph{FaSTGLZ} algorithm in Section \ref{sec:fastglz_alg}, where we also present pseudo-code that details the main algorithmic steps.  We then validate both the computational efficiency and usefulness of our algorithm by applying it to real-world datasets in Section \ref{sec:results}, and conclude in Section \ref{sec:conclusion}.  

\section{Preliminaries}
\label{sec:prelim}

We start with a dataset $\left\{(x^{(i)},y^{(i)})\right\}_{i=1}^n$, with features $x^{(i)}\in\R^p$ and response $y^{(i)}$.  Depending on the application, $y^{(i)}$ may be a continuous value or a categorical label.  For convenience, the feature data will be assembled into a $p\times n$ data matrix $X$.  
%In the context of fMRI, the features may represent the BOLD responses from brain voxels, while  $y$ may indicate the category of the presented stimulus.  

Many machine learning algorithms attempt to predict the response $y$ from some linear projection of the data, $\eta(w)=w^Tx$, where $w\in\R^p$ weights the relative importance of the features.  Given a loss function $L(\eta(w),y)$ that measures the fidelity of the prediction, $w$ is estimated by minimizing:
\begin{equation}
w^*=\arg\min_w \sum_{i=1}^n d^{(i)} L(\eta^{(i)}(w),y^{(i)}) + \lambda R(w) \label{eqn:general_problem}
\end{equation}
where $d^{(i)}$ weights the importance of the $i^{th}$ trial on the optimization and $R(w)$ is a regularization term that reduces over-fitting to the training data and/or improves the interpretability of the model.  In this paper, we develop a fast algorithm for solving (\ref{eqn:general_problem}) when the loss function derives from the negative log-likelihood of a generalized linear model (GLZ), and the regularization is the elastic net penalty.  Before outlining our \emph{FaSTGLZ} algorithm in Section \ref{sec:fastglz_alg}, the remainder of this section provides a brief introduction to GLZ's and the elastic net penalty, while also framing the main problem that we seek to solve.  We also discuss how our algorithm may be extended to other loss functions and regularizers, such as the group lasso, in Section \ref{sec:extensions_fastglz}.

GLZ's assume that the conditional distribution of $y$ given the data $x$ is a member of the exponential family of distributions:
\begin{equation}
p(y|x) = \exp\left(\frac{y\eta - b(\eta)}{a(\phi)}+c(y,\phi)\right)
\label{eqn:glz}
\end{equation}
where $\eta=x^Tw$ is, again, a linear predictor and $\phi$ is a dispersion parameter.  The functions $a,b,c$ fully specify the distribution from the  family.  Table \ref{table:glzs} lists a few of the common regression models along with the associated definitions for $a,b$, and $c$.  For more information on GLZ's, see \citet{McCullagh89}.
\begin{table}[H]
\begin{center}
\begin{tabular}{|c|c|c|c|}
\hline
Regression Model	&	$a(\phi)$			&	$b(\eta)$				&	$c(y,\phi)$	\\
\hline
Linear Regression	&	$\phi^2$			&	$\frac{1}{2}\eta^2$		&	$\frac{y^2}{2\phi^2}$			\\
\hline
Logistic Regression	&	$1$				&	$\log(1+\exp(\eta))$		&	$0$			\\
\hline
Poisson Regression	&	$1$				&	$\exp(\eta)$			&	$-\log(y!)$			\\
\hline
\end{tabular}
\caption{A listing of common GLZ's, with their associated settings for functions $a(\phi)$, $b(\eta)$, and $c(y,\phi)$ in the conditional distribution $p(y|x)$ of (\ref{eqn:glz}).}
\label{table:glzs}
\end{center}
\end{table}

The conditional mean and variance of $y$, written as $\mu(w)$ and $\Var(w)$ to emphasize their dependence on the feature weights $w$, are given by:
\begin{eqnarray}
\mu(w) &=& \frac{db(\eta)}{d\eta}|_{\eta=x^Tw} \\
\Var(w) &=& a(\phi)\frac{d^2b(\eta)}{d\eta^2}|_{\eta=x^Tw} 
\end{eqnarray}
The inverse of the mean function $\mu(w)$ is often referred to as the link function, as it relates the mean of the dependent variable $y$ to the linear predictor $\eta$.

Given a data sample $\{x^{(i)},y^{(i)}\}_{i=1}^n$ and associated trial weightings $d^{(1)},\dots,d^{(n)}$, we may estimate $w$ by minimizing a regularized negative log-likelihood:
\begin{equation}
J(w) = \ell(w) + \lambda_1||w||_1 + \lambda_2||w||_2^2 \label{eqn:elnet_single_prob}
\end{equation}
where $\lambda_1,\lambda_2\geq 0$ are tuning parameters, and $\ell(w)$ is given by:
\begin{equation}
\ell(w) = -\sum_{i=1}^n d^{(i)} \left[y^{(i)}\eta^{(i)}(w) - b(\eta^{(i)}(w))\right] \label{eqn:ell_single_prob}
\end{equation}
For simplicity, we assume that the dispersion parameters $\phi^{(1)},\dots,\phi^{(n)}$ are known, and the $1/a(\phi^{(i)})$ term has been absorbed into $d^{(i)}$.

Minimizing (\ref{eqn:elnet_single_prob}) is a convex optimization problem, for which many efficient algorithms have been proposed \citep[e.g.][]{Friedman10}.  However, our goal is to simultaneously solve a multitude of such problems that are derived from the same dataset.  Since each problem will generally optimize $J(w)$ with respect to a distinct version of the data, each will have its own log-likelihood term $\ell_k(w_k)$, where $w_k$ represents the unknown weights for problem $k\in\{1,\dots,K\}$.  For clarity, we use a subscript $k$ on a variable to emphasize that it is specific to the $k^{th}$ problem. The allowable variability in $\ell_k(w_k)$ may be expressed by introducing problem-specific trial weighting vector $d_k$ and response vector $y_k$, so that (\ref{eqn:ell_single_prob}) is adapted to:
\begin{equation}
\ell_k(w_k) = -\sum_{i=1}^n d_k^{(i)}\left[y_k^{(i)}\eta^{(i)}(w_k) - b(\eta^{(i)}(w_k))\right] \label{eqn:ell_all_probs}
\end{equation}

Cross-validation, bootstrapping, and nonparametric significance testing all fall under this framework.  For example, let $F_k=\left[f_{k1},\dots,f_{kn}\right]$ denote the relative frequencies of the training examples derived from a bootstrap or cross-validation fold.  Its log-likelihood $\ell_k(w_k)$ may be expressed in the form of (\ref{eqn:ell_all_probs}) by setting $d_k^{(i)}$ to $f_{ki}$.  Note that if the $j^{th}$ sample is excluded (e.g., a sample belonging to the validation set of a cross-validation fold), then $d_k^{(j)}=0$ and the $j^{th}$ sample does not exert any influence on the objective.  

Significance testing by nonparametric permutation testing \citep{Golland05} also fits the form of (\ref{eqn:ell_all_probs}).  Here, the GLZ is re-trained on new datasets in which the response $y$ has been permuted across examples.  In this case, each problem $k$ is given its own $y_k$, which is a permutation of the original sample.

\begin{figure}[H]
\vskip -0.1in
\centering
\subfigure[Example dataset with $n=5$ trials, color-coded for visualization purposes.]{
	\includegraphics[width=0.75\linewidth]{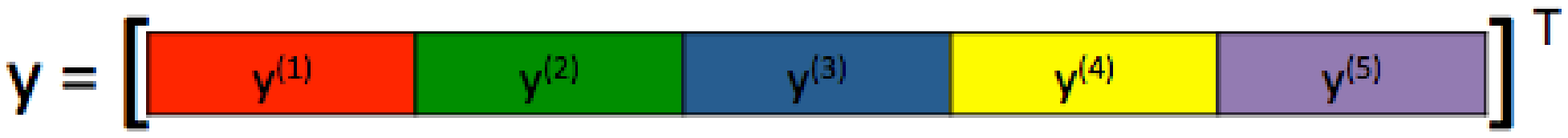}
	\label{fig:relspeed_against_1e-4}
}
\subfigure[Problem structure corresponding to leave-one-out cross-validation ($K=n$ problems). Corresponding columns of $D$ and $Y$ define a single optimization problem. In the trial weighting matrix $D$, training set trials are weighted by $1/4$ and cells shaded in gray represent trials that belong to the test set (with a weight of $0$), and hence do not influence the optimization. The $K$ optimization problems share the same response vector $y$.]{
	\includegraphics[width=\linewidth]{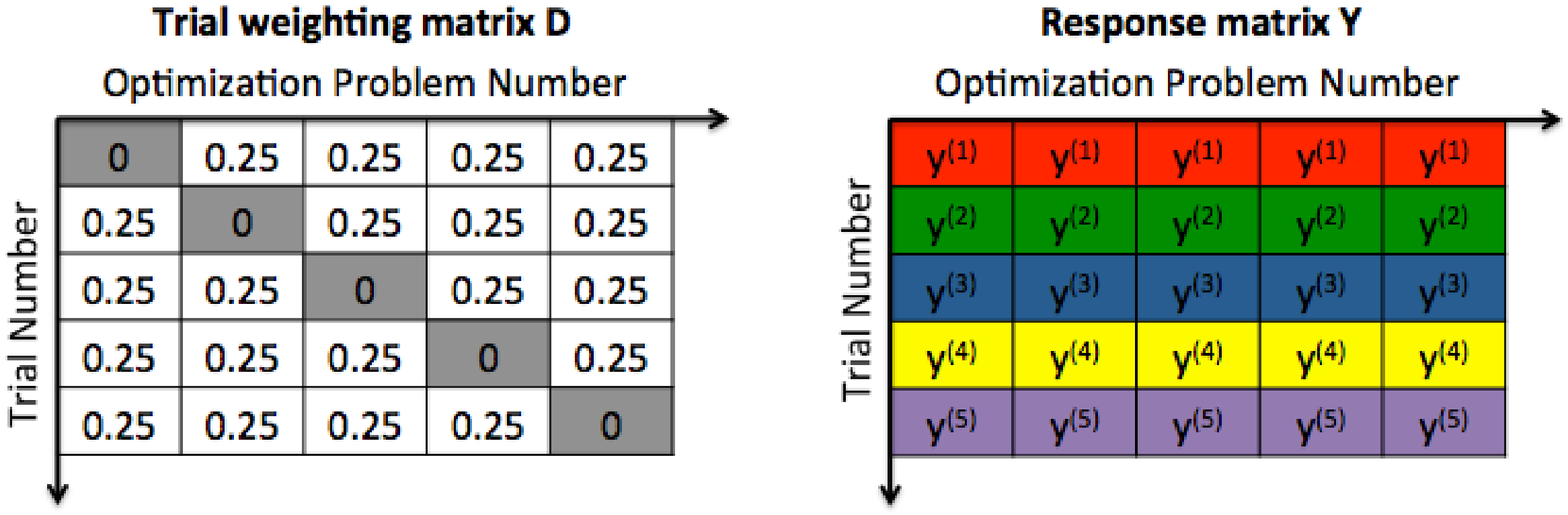}
	\label{fig:objval_against_1e-4}
}
\subfigure[Permutation testing and cross-validation can be combined by varying $D$ and $Y$ together. This example problem structure performs leave-one-out cross-validation for each of $m$ permutations ($K=mn$ total problems).]{
	\includegraphics[width=\linewidth]{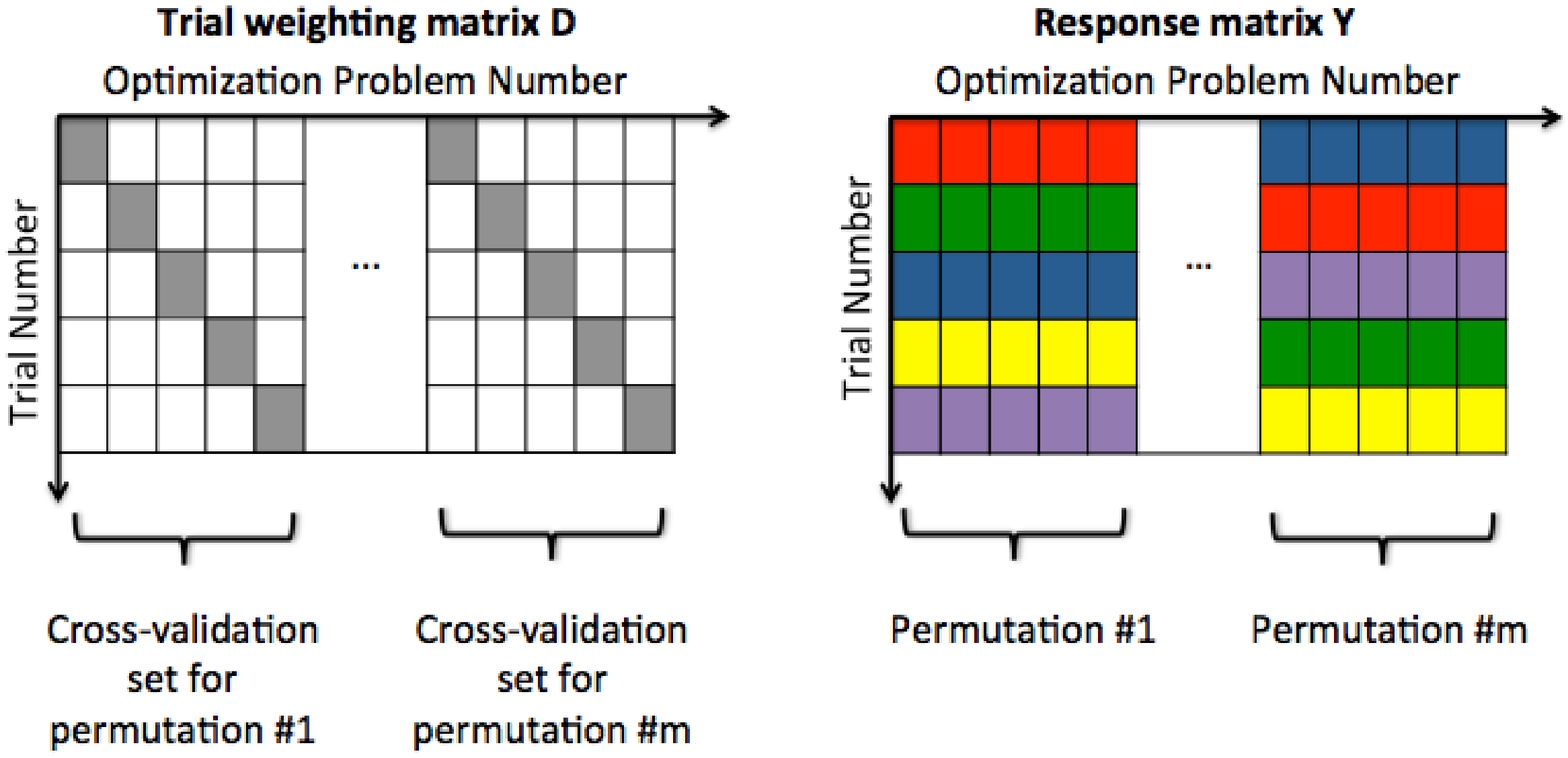}
	\label{fig:objval_against_1e-4}
}
\vskip -0.1in
\caption{Illustrative optimization problem structures that can be solved simultaneously by \emph{FaSTGLZ}. Corresponding columns of the trial weighting matrix $D$ and response matrix $Y$ define a single optimization problem.}
\label{fig:xvalandperm}
\end{figure}

To summarize, we seek to minimize the objectives $J_k(w_k)$, $k=1,\dots,K$:
\begin{equation}
\min_{w_k} J_k(w_k) = \min_{w_k} \ell_k(w_k) + \lambda_1||w_k||_1 + \lambda_2||w_k||_2^2
\label{eqn:J_all_probs}
\end{equation}
Under cross-validation and bootstrapping, the variability in $\ell_k$ arises through problem-specific $d_k$, while permutation testing utilizes distinct $y_k$.  Thus, we may characterize the entire set of problems arising from any combination of cross-validation, bootstrapping, and permutation testing by $n\times K$ matrices $D$ and $Y$, in which the $k^{th}$ columns of $D$ and $Y$ contain the trial weighting vector $d_k$ and response vector $y_k$ that correspond to the $k^{th}$ problem.  Example problem structures are illustrated in Figure \ref{fig:xvalandperm}.

\begin{comment}
\textbf{Here show that all problems can be characterized by a matrix $D$, $Y$, and weight matrix $W$.  Talk about size of $W$ which would normally limit problem size, but this is handled in a later section.  In fact, we show that $W$ can be stored with $O(nK)$ if $\lambda_1=0$, and $O(nK+s)$ if $\lambda_1\neq 0$ where $s=$}.
\end{comment}

\section{The FaSTGLZ Algorithm}
\label{sec:fastglz_alg}

This section presents the \emph{FaSTGLZ} algorithm.  After introducing the major components of the algorithm in Sections \ref{sec:simult_newt_meth} and \ref{sec:fastglz_with_sparsity}, we offer a discussion on the algorithm's memory overhead in Section \ref{subsec:screen_rules}, as well as extensions of the algorithm to other loss functions and regularizers, including the group lasso in Section \ref{sec:extensions_fastglz}.

\subsection{Simultaneous Newton Solver}
\label{sec:simult_newt_meth}

We first consider simultaneously solving (\ref{eqn:J_all_probs}) without the sparsity constraint, by setting $\lambda_1=0$.  This greatly simplifies the problem by removing the non-differentiable portion of the objective. It also allows us to highlight our simultaneous Newton solver, which is a major component of the complete \emph{FaSTGLZ} algorithm, and is the main machinery through which the shared structure across problems is leveraged for computational efficiency.  

With $\lambda_1=0$, our goal reduces to minimizing the $K$ objectives $J_k(w_k)$:
\begin{equation}
J_k(w_k) = \ell_k(w_k) + \lambda_2||w_k||_2^2 \label{eqn:J_k_without_sparsity}
\end{equation}
This amounts to fitting a GLZ with a ridge penalty.  This is commonly optimized by iteratively re-weighted least squares (IRLS), which sequentially minimizes a quadratic approximation to (\ref{eqn:J_k_without_sparsity}).  Specifically, $\ell_k(w_k)$ is approximated by a quadratic function $q_k(w_k,\bar w_k)$ around an estimate $\bar w_k$:
\begin{equation}
q_k(w_k,\bar w_k) = \ell_k(\bar w_k) + \nabla\ell_k^T(w_k-\bar w_k) + \frac{1}{2}(w_k-\bar w_k)^TH_k(w_k-\bar w_k) \label{eqn:quad_approx}
\end{equation}
\vskip -2em
\begin{eqnarray}
\nabla\ell_k &=& Xe_k \label{eqn:ell_grad} \\
H_k &=& XR_kX^T \label{eqn:ell_hess}
\end{eqnarray}
where $\nabla\ell_k$ and $H_k$ are, respectively, the gradient and Hessian of $\ell_k(w_k)$ evaluated at $\bar w_k$, $e_k=d_k\circ (\mu(\bar w_k)-y_k)$ is the residual error vector and $\circ$ is the Hadamard product. Also, $R_k$ is a diagonal matrix of non-negative values, whose $i^{th}$ diagonal entry is given by $d_k^{(i)}\Var^{(i)}(\bar w_k)$.

Since the Hessian $H_k$ is positive semi-definite, we may solve for the minimum of the quadratic approximation to $J_k(w_k)$ by setting its gradient to zero.  This results in the following system of linear equations:
\begin{equation}
(H_k + 2\lambda_2 I)w_k = Xb_k \label{eqn:linear_sys}
\end{equation}
where $b_k=R_k\eta(\bar w_k)-e_k$.  Inverting this linear system, either directly or by an iterative method, would typically be prohibitive since the number of features $p$ is assumed to be large and $H_k$ is not a sparse matrix.

At this point, we take advantage of the shared structure in two key ways in order to simultaneously solve (\ref{eqn:linear_sys}) across all $K$ problems very efficiently. First, analogous to the representer theorem, as described in \citet{Kimeldorf70,Scholkopf01}, we note that the solution $w_k$ in (\ref{eqn:linear_sys}) must lie in the range (column space) of the data matrix $X$ for all $k=1,\dots,K$.  This allows us to transform the linear system above into a much lower-dimensional space.  Specifically, given a QR-factorization for the data matrix, $X=QZ$, with $Q\in\R^{p\times n}$ having orthonormal columns and $Z\in\R^{n\times n}$, the solution may be expressed as $w_k=Q\alpha_k$, where $\alpha_k$ satisfies:
\begin{eqnarray}
G_k\alpha_k &=& Zb_k\text{, } \label{eqn:linear_sys_reduced} \\
G_k &=& ZR_kZ^T + 2\lambda_2 I \label{eqn:G_k}
\end{eqnarray}
Thus, we have converted the original linear system of $p$ equations into a system of $n$ equations by projecting into a common low-dimensional space spanned by the columns of $Q$.  In $p\gg n$ scenarios, this results in a significant reduction.
%Particularly in the case of fMRI, this results in a significant reduction.  Whereas the number of features (voxels) $p$ can be in the tens of thousands, the number of trials $n$ is typically limited to at most $1000$ due to the temporal coarseness of the functional image sampling, fixed by the repetition time TR.

The second way that we exploit the shared structure allows us to solve (\ref{eqn:linear_sys_reduced}) for all $K$ problems with only a single matrix inversion and in one linear algebraic expression.  Before deriving the method, we first briefly summarize the main ideas.  Since the structural form of $G_k$ is highly similar across problems, we first approximate the solution to (\ref{eqn:linear_sys_reduced}) by inverting a template matrix $M$ that is representative of all the $G_k$.  To obtain the exact solution for each $k$, we then employ an iterative algorithm that corrects the errors incurred by approximating each $G_k$ by the template $M$.  Since the variability between $G_k$ and $M$ arises through the matrix $R_k$ in (\ref{eqn:G_k}), it cannot be modeled as a low-rank perturbation, for which numerous correction methods exist \citep[see][]{Golub96}.  Instead, we base our iterative correction algorithm on the theory of stationary iterative methods for solving linear systems of equations \citep{Young71}.

Stationary iterative methods provide a simple and efficient framework for solving linear systems $Gx=b$ without resorting to inverting $G$.  Instead, $G$ is decomposed into an additive splitting of two matrices $G=M-N$, where the inverse of $M$ is known or easily computable, and $N$ is the residual.  The linear system $Gx=b$ is then solved by computing a sequence of iterates:
\begin{equation}
x^{(t+1)} = M^{-1}Nx^{(t)} + M^{-1}b \label{eqn:stationary_iterative_eqn}
\end{equation}
Convergence to the solution $x=G^{-1}b$ is guaranteed as long as the spectral radius of $M^{-1}N$ is less than one \citep{Young71}.  Intuitively, this provision requires that the template $M$ sufficiently resembles $G$.  

Rather than using this technique to invert a single matrix, we will apply it to invert all $G_k$, $k=1,\dots,K$. To do so, we define a common template matrix $M=ZRZ^T+2\lambda_2 I$ whose inverse we will compute, where $R$ is a diagonal matrix of non-negative values that we will specify shortly.  We can then express each $G_k$ in terms of the template plus a residual:
\begin{equation}
G_k = M - Z(R-R_k)Z^T
\end{equation}
Using (\ref{eqn:stationary_iterative_eqn}), this splitting implies the following iteration for solving (\ref{eqn:linear_sys_reduced}):
\begin{equation}
\alpha_k^{(t+1)} = M^{-1}Z(R-R_k)Z^T\alpha_k^{(t)} + M^{-1}Zb_k \label{eqn:simult_main_iteration}
\end{equation}
We prove in the supplementary material that taking the $n\times n$ matrix $R$ to be the element-wise maximum $R=\max(R_1,\dots,R_K)$ guarantees that $\alpha^{(t+1)}$ converges to the true solution of (\ref{eqn:linear_sys_reduced}) for every $k$. See also \citet{Conroy12}.

An added benefit of the iteration in (\ref{eqn:simult_main_iteration}) is that the updates across $k$ can be pooled into a single linear algebraic expression.  This greatly simplifies the code and also minimizes memory access times.  Let $A^{(t)},B,R_{\Delta}$ be $n\times K$ matrices whose $k^{th}$ columns contain, respectively, $\alpha_k^{(t)}$, $b_k$ and $\text{diag}(R-R_k)$, where $\text{diag}(\cdot)$ extracts the main diagonal entries of a matrix into a vector.  Then the updates in (\ref{eqn:simult_main_iteration}) can be computed as:
\begin{equation}
A^{(t+1)} = M^{-1}Z\left[R_{\Delta}\circ (Z^TA^{(t)})\right] + M^{-1}ZB \label{eqn:iteration_one_algebraic}
\end{equation}
Although computing $M^{-1}Z$ requires $O(n^3)$ operations, this is only performed once at initialization.  Otherwise, the complexity of the iteration above is $O(n^2K)$.  Thus, computing Newton's method in this manner scales with the number of problems $K$ like $O(n^2K)$ instead of $O(n^3K)$ (ignoring terms independent of $K$).  Also note that the number of iterations required for the sequence in (\ref{eqn:iteration_one_algebraic}) to converge depends entirely on how well the template matches the true $G_k$ matrices, and is not a function of $n$.  

Minimization then proceeds by iteratively updating the quadratic approximation in (\ref{eqn:quad_approx}), each time setting $\bar w_k$ to the previously estimated $w_k$. Since Newton's method has local quadratic convergence properties \citep{Dennis87}, very few iterations are required in practice.

\subsubsection{Example of Broader Utility of Simultaneous Newton Solver}

Linear systems of the form in (\ref{eqn:linear_sys}) arise in a wide array of situations, particularly in optimization problems that may be solved by iteratively re-weighted least squares. Here we highlight the broader utility of our simultaneous Newton solver through an example in time-series regression. Suppose we have a set of $K$ time-series $y_k$, $k=1,\dots,K$ that we would like to linearly regress against a common set of signals, assembled as columns in the matrix $X$:
\begin{equation}
y_k = X w_k + \epsilon_k
\end{equation}
where $\epsilon_k \sim {\cal{N}}(0,V_k)$ is colored Gaussian noise whose temporal auto-correlation structure $V_k$ varies with $k$.  Such a model, for example, arises in the univariate General Linear Model in functional magnetic resonance imaging (fMRI) statistical analysis \citep{Woolrich01}.  In this case, the time-series are the measured fMRI signals from a set of $K$ brain voxels, and $X$ is a design matrix that encodes information about the experimental conditions and confounds. Having the noise auto-covariance depend on $k$ allows the model to adapt to the spatially-varying noise properties across the brain.

Given the model above, the best linear unbiased estimator (BLUE) minimizes the negative log-likelihood:
\begin{equation}
w_k^* = \arg\min_{w_k} (y_k-Xw_k)^TV_k^{-1}(y_k-Xw_k) \label{eqn:tsr_blue}
\end{equation}
A common assumption in time-series regression is that the noise is generated by an auto-regressive process, in which case there exists a Toeplitz matrix $S_k$ such that $V_k^{-1}=S_k^TS_k$.  In fact, $S_k$ implements the filter that whitens the noise, and may be estimated by analyzing the covariance of the initial residuals $(y_k-X\hat w_k)$, with $\hat w_k$ minimizing (\ref{eqn:tsr_blue}) when $V_k=I$.  Given an estimate for $S_k$, our original problem (\ref{eqn:tsr_blue}) reduces to:
\begin{eqnarray}
w_k^* &=& \arg\min_{w_k} ||S_ky_k - S_kXw_k||^2 \\
	&=& (X^TS_k^TS_kX)^{-1}X^TS_k^TS_ky_k \label{eqn:tsr_whitened_sol}
\end{eqnarray}
This process is repeated for each time-series regression problem $k=1,\dots,K$.

The solution to (\ref{eqn:tsr_whitened_sol}) can be implemented entirely in the frequency domain and fits naturally into the simultaneous Newton framework proposed above.  This approach avoids computing multiple inverse Fourier transforms and matrix inverses.  With sufficient zero-padding, the $S_k$ matrices can be made to be circulant, in which case they share a common set of eigenvectors that correspond to the DFT basis.  Specifically, $S_k$ can be decomposed as $S_k=UR_kU^H$, where $R_k$ is a diagonal matrix that contains the DFT coefficients of the $k^{th}$ auto-correlation filter, and $U^H$ is the unitary DFT matrix so that $\hat v=U^Hv$ produces the DFT coefficients of a signal $v$.  Substituting this expression for $S_k$ into (\ref{eqn:tsr_whitened_sol}), we obtain:
\begin{equation}
w_k^* = (\hat X^H R_k^HR_k \hat X)^{-1}\hat X^H R_k^HR_k \hat y_k \label{eqn:tsr_in_freq_dom}
\end{equation}
where $\hat X$ and $\hat y_k$ are the DFT coefficients of the design matrix regressors and response time-series, respectively.  Since $R_k^HR_k$ is a real-valued diagonal matrix, the matrix inversions in (\ref{eqn:tsr_in_freq_dom}) may be computed simultaneously as before, with the template matrix taking the form of $M=\hat X^H R \hat X$, and $R$ is a diagonal matrix whose entries are the element-wise maxima of $R_1^HR_1,\dots,R_K^HR_K$.

\subsection{FaSTGLZ with sparsity}
\label{sec:fastglz_with_sparsity}

We now return to the main problem (\ref{eqn:J_all_probs}), and this time also consider the sparsity-inducing term $||w_k||_1$.  In this setting, we base our approach on the optimization framework of alternating direction method of multipliers (ADMM) \citep{Eckstein92}.  The reasoning for this decision is two-fold:  first, ADMM provides a natural way for us to employ our simultaneous Newton solver that exploits the redundant problem structure; and second, ADMM is flexible enough to accommodate other regularizers, including the group lasso.  We explore this extension further in Section \ref{sec:extensions_fastglz}.

For each problem $k=1,\dots,K$, we divide the objective function $J_k(w_k)$ in (\ref{eqn:J_all_probs}) into the sum of two terms: the differentiable portion $f_k(w_k)=\ell_k(w_k)+\lambda_2||w_k||_2^2$, and the non-differentiable $\ell_1$ term $g(w_k)=\lambda_1||w_k||_1$.  ADMM then formulates an equivalent optimization problem through the addition of an auxiliary variable $v_k\in\R^p$:
\begin{eqnarray*}
\min_{w_k,v_k} \ell_k(w_k) + \lambda_2 ||w_k||_2^2 + \lambda_1 ||v_k||_1 \\
\text{subject to } w_k=v_k
\end{eqnarray*}
whose augmented Lagrangian may be expressed as:
\begin{equation}
{\cal{L}}_k(w_k,v_k) = f_k(w_k) + g(v_k) - \lambda_k^T(w_k-v_k) + \frac{1}{2\mu}||w_k-v_k||_2^2
\label{eqn:augmented_lagrangian}
\end{equation}
where $\lambda_k\in\R^p$ are estimates of the Lagrange multipliers and $\mu\geq 0$ is a penalty parameter. A benefit of minimizing the augmented Lagrangian is that the constraint $w_k=v_k$ can be satisfied without taking $\mu\rightarrow 0$ \citep{Afonso10}.

Optimization proceeds by alternating between minimizing (\ref{eqn:augmented_lagrangian}) with respect to $w_k$ while holding $v_k$ fixed, and vice versa.  This is equivalent to the symmetric alternating direction augmented Lagrangian method (SADAL) described in \citet{Goldfarb09}.  Specifically, the algorithmic steps are:
\begin{eqnarray}
w_k &\leftarrow& \arg\min_{w_k} {\cal{L}}_k(w_k,v_k) \label{eqn:admm_step1} \\
\lambda_k &\leftarrow& \lambda_k - (1/\mu)(w_k-v_k) \label{eqn:admm_step2} \\
v_k &\leftarrow& \arg\min_{v_k} {\cal{L}}_k(w_k,v_k) \label{eqn:admm_step3} \\
\lambda_k &\leftarrow& \lambda_k - (1/\mu)(w_k-v_k) \label{eqn:admm_step4}
\end{eqnarray}
The resulting subproblems are substantially simpler than the original: optimizing $w_k$ in (\ref{eqn:admm_step1}) involves a differentiable objective, while updating $v_k$ in (\ref{eqn:admm_step3}) reduces to a soft-thresholding operation.  

For our purposes, it is more convenient to re-parameterize the algorithmic steps in terms of a variable $l_k$, which is related to the Lagrange multiplier estimates $\lambda_k$.  
%The algorithmic steps for solving the above follow analogously as in those presented in Section \ref{sec:component_admm}.  However, to simplify notation, it is more convenient to pose the steps in an equivalent form given below (see Appendix for derivation).  
Specifically, given initial values for $w_k,v_k,\lambda_k$, we initialize $l_k$ to $l_k=\lambda_k+(1/\mu)v_k$.  Then the ADMM procedure is equivalent to:
\begin{eqnarray}
w_k &\leftarrow& \arg\min_{w_k} S_k(w_k,l_k) \label{eqn:min_Sk} \\
l_k &\leftarrow& l_k - (2/\mu)w_k \label{eqn:admmmod_step2} \\
v_k &\leftarrow& -\mu\text{ soft}(l_k,\lambda_1\textbf{1}) \label{eqn:admmmod_step3} \\
l_k &\leftarrow& l_k + (2/\mu)v_k \label{eqn:admmmod_step4}
\end{eqnarray}
where
\begin{eqnarray}
S_k(w_k,l_k) &=& \ell_k(w_k) + \rho||w_k||_2^2 - l_k^Tw_k \label{eqn:Sk}
\end{eqnarray}
and $\rho = \lambda_2 + \frac{1}{2\mu}$. Also, $\text{soft}(a,b)=\text{sgn}(a)\max(|a|-b,0)$ is the soft-thresholding operator. The equivalence between (\ref{eqn:admm_step1})-(\ref{eqn:admm_step4}) and (\ref{eqn:min_Sk})-(\ref{eqn:admmmod_step4}) is shown in the supplementary material.
 
Note that the updates in (\ref{eqn:admmmod_step2}-\ref{eqn:admmmod_step4}) are computable in closed-form and the only remaining challenge is minimizing the smooth function $S_k(w_k,l_k)$ in (\ref{eqn:min_Sk}).  Minimizing $S_k(w_k,l_k)$, however, is almost completely analogous to the ridge-penalized GLZ in (\ref{eqn:J_k_without_sparsity}), except for the extra linear term $l_k^Tw_k$.  This allows us to apply our simultaneous Newton's method solver to (\ref{eqn:min_Sk}) with only a slight modification. As before, we sequentially minimize $S_k(w_k,l_k)$ by taking a quadratic approximation to $\ell_k(w_k)$ around an initial estimate $\bar w_k$.  Then, given a QR-factorization of the data matrix $X=QZ$, we prove in the supplementary material that the minimum to the quadratic approximation to $S_k(w_k,l_k)$ is attained at:
\begin{equation}
w_k = Q\alpha_k + \frac{1}{2\rho}(I-P_Q)l_k \label{eqn:wk}
\end{equation}
where $P_Q$ is the projection onto $\text{range}(Q)$, and $\alpha_k\in\R^n$ satisfies:
\begin{equation}
(ZR_kZ^T+2\rho I)\alpha_k=Zb_k+Q^Tl_k \label{eqn:sub_linear_system}
\end{equation}
The $\alpha_k$ are then computed simultaneously for all $k$ using the techniques outlined in Section \ref{sec:simult_newt_meth} for solving (\ref{eqn:linear_sys_reduced}).  As before, (\ref{eqn:sub_linear_system}) can be solved for all $k$ with only one matrix inversion and in one linear algebraic expression.

 \subsection{Reducing memory overhead}
 \label{subsec:screen_rules}
 
As formulated thus far, $l_k,w_k,v_k$ must be stored in full for each $k$, which requires $O(pK)$ memory elements.  Since $p$ is often very large, this places a practical constraint on the number of problems $K$ that may be solved simultaneously.  This memory overhead can be significantly reduced by employing the recently proposed screening rules for $\ell_1$-regularized problems to estimate the active set of each problem \citep{Tibshirani12}.  These tests are based on correlations between the features and the response variable, and can thus be evaluated very efficiently.  

Given an estimate of the active set $A_k$ for each $k=1,\dots,K$, we soft-threshold in (\ref{eqn:admmmod_step3}) only to the active set $A_k$.  Thus, $v_k$ is a sparse vector.  Additionally, we show here that the algorithmic steps in (\ref{eqn:min_Sk})-(\ref{eqn:admmmod_step4}) may be computed by only storing $\alpha_k=Q^Tw_k$, $\beta_k=Q^Tl_k$, and the sparse set of entries of $l_k$ restricted to $A_k$.  This is an important point because even though the support of $v_k$ is always confined to the current estimate of the active set, $w_k$ and $l_k$ are not necessarily sparse vectors over all iterations.  So our memory overhead has been reduced to $O((n+s)K)$, where $s=\sum_{k=1}^K|A_k|/K$.  Since $n\ll p$, this is more scalable for moderate sparsity levels.

To illustrate, assume we are given initial conditions for $\beta_k$, and the elements of $l_k$ belonging to $A_k$.  Upon minimizing (\ref{eqn:min_Sk}) by solving for $\alpha_k$ from the linear system in (\ref{eqn:sub_linear_system}), we must update both $l_k$ and $\beta_k$ through (\ref{eqn:admmmod_step2}).  Using the expression for $w_k$ in (\ref{eqn:wk}), this can be computed as:
\begin{eqnarray}
l_k &\leftarrow& l_k-(2/\mu)w_k \\
	&\leftarrow& \left(1-\frac{1}{\mu\rho}\right)l_k + \frac{2}{\mu}Q\left(\frac{1}{2\rho}\beta_k-\alpha_k\right) \label{eqn:lk_update_1} \\
\beta_k &\leftarrow& \beta_k - (2/\mu)\alpha_k
\end{eqnarray}

Upon updating $v_k$ by soft-thresholding the entries of $l_k$ belonging to $A_k$, the second update of $l_k$ in (\ref{eqn:admmmod_step4}) is straightforward:
\begin{eqnarray}
l_k &\leftarrow& l_k + (2/\mu)v_k \label{eqn:lk_update_2} \\
\beta_k &\leftarrow& \beta_k + (2/\mu)Q^Tv_k \label{eqn:bk_update_2}
\end{eqnarray}

The above updates are computed using only $\beta_k,\alpha_k$, and $v_k$.  Also, since the soft-thresholding step only requires the elements of $l_k$ at $A_k$, and the updates for each element of $l_k$ in (\ref{eqn:lk_update_1}) and (\ref{eqn:lk_update_2}) do not depend on any other elements of $l_k$, we only store the elements of $l_k$ belonging to $A_k$.  Aside from the reduced memory load, this insight also improves the computational complexity of these updates.  Specifically, the matrix multiplication by $Q$ in (\ref{eqn:lk_update_1}) must only be computed amongst the rows of $Q$ that belong to $A_k$.  This reduces the number of multiplications to $O(snK)$ from $O(pnK)$.  The other multiplication by $Q^T$ in (\ref{eqn:bk_update_2}) also requires only $O(snK)$ multiplications.

We can make the memory constraint explicit by estimating the maximum amount of memory required as a function of $p,n,s_{\max}$ and $K$, where $s_{\max}$ is the maximum allowable number of features to include in each model.  Since sparsity is desirable, $s_{\max}$ is usually capped at a small fraction of the total number of features.  The amount of memory, in units of bytes, is given by:
\begin{eqnarray*}
\text{Memory} &=& \left(64s_{\max}+40n+32\min(p,n)+40\right)K \\
			&+& 8pn + 24n\min(p,n) + 16\min(p,n)^2\text{ (bytes)}
\end{eqnarray*}
Assuming that $n\ll p$, the amount of memory required grows with the number of problems $K$ at a rate roughly proportional to $(n+s_{\text{max}})K$.  To better contextualize this, consider a machine learning dataset with $p=50,000$ features and $n=500$ trials. Capping the number of allowable features included in the model at $s_{\max}=1,000$, the amount of memory required in megabytes as a function of $K$ is $(200+0.1K)$MB.  Thus, more than $8,000$ problems can be solved simultaneously with $1$GB of RAM.

\begin{comment}
The steps of the \emph{FaSTGLZ} algorithm are summarized in Algorithm \ref{alg:alg1}, and 
\end{comment}
Table \ref{table:variable_desc} provides a description of the main variables used by the algorithm.

\begin{table}[H]
\begin{center}
\begin{tabular}{|c|c|l|}
\hline
Variable	&	Dimension	&	Description \\
\hline
$p$		&	Scalar		&	The feature space dimension (number of voxels) \\
\hline
$n$		& 	Scalar		&	The number of trials \\
\hline
$K$		&	Scalar		&	The total number of problems to be solved for \\
\hline
$X$		&	$p\times n$	&	The data matrix (features x trials) \\
\hline
$Q$		&	$p\times n$	&	Orthonormal basis for the range of $X$ \\
\hline
$Z$		&	$n\times n$	&	$Z=Q^TX$ \\
\hline
$d_k$	&	$n\times 1$	&	The trial weighting vector for the $k^{th}$ problem \\
\hline
$y_k$	&	$n\times 1$	&	The response vector to predict for the $k^{th}$ problem \\
\hline
$w_k,v_k$	&	$p\times 1$	&	Discriminative weights for the $k^{th}$ problem \\
\hline
$l_k$	&	$p\times 1$	&	Lagrange multipliers for ADMM procedure \\
\hline
\end{tabular}
\caption{A description of the major variables involved in the FaSTGLZ algorithm.}
\label{table:variable_desc}
\end{center}
\vskip -1.2em
\end{table}

\subsection{Extensions of the FaSTGLZ Algorithm}
\label{sec:extensions_fastglz}

In this section we discuss extensions of our \emph{FaSTGLZ} algorithm to solving (\ref{eqn:general_problem}) for other loss functions and regularizers.  We first discuss regularizers other than just the elastic net, and then move to loss functions other than those derived from GLZ's.

As mentioned previously, ADMM is a flexible technique that can accommodate regularizers other than just the elastic net.  Let $R(w)=\lambda_2||w||_2^2 + g(w)$ be the regularization penalty, where $g(w)$ is some possibly non-differentiable penalty function. Employing the ADDM splitting technique as before, we set $f(w)=\ell(w)+\lambda_2||w||_2^2$ and obtain the following augmented Lagrangian:
\begin{equation}
{\cal{L}}_k(w_k,v_k) = f_k(w_k) + g(v_k) - \lambda_k^T(w_k-v_k) + \frac{1}{2\mu}||w_k-v_k||_2^2
\end{equation}
In pursuing the alternating minimization strategy as in (\ref{eqn:min_Sk}-\ref{eqn:admmmod_step4}), only the minimization with respect to $v_k$ differs from before.  Specifically, we end up with the following update:
\begin{equation}
v_k \leftarrow \arg\min_{v_k} g(v_k) + \frac{1}{2\mu}\left\|\mu l_k + v_k\right\|_2^2 \label{eqn:proximity_operator}
\end{equation}
where, as before, $l_k=\lambda_k-(1/\mu)w_k$.  This minimization is often called the proximity operator of the function $g$ \citep{Boyd10}. It turns out that the proximity operator has a closed-form solution for many useful and popular regularizers.  For example, when $g(v_k)=||v_k||_1$ as in the elastic net, the update results in a soft-thresholding operation.

The group lasso \citep{Yuan07,Meier08} is another example whose proximity operator may be computed in closed-form. The group lasso is useful when prior information allows one to cluster features into distinct groups.  The group lasso penalty then encourages sparsity across groups but not within a group, so that features from a group are either all included or all excluded.  This penalty may arise in fMRI, for example, given a parcellation that clusters voxels into pre-defined regions-of-interest (ROI).  The group lasso would then encourage sparsity across distinct ROI's, but all voxels in a given ROI selected by the model would contribute.  Given a grouping of the $p$ features, with ${\cal{I}}_i$ denoting the index set of the features belonging to the $i^{th}$ group, the group lasso penalty is defined as:
\begin{equation}
g(w) = \lambda_1\sum_i ||w_{{\cal{I}}_i}||_2
\end{equation}
Plugging this penalty into (\ref{eqn:proximity_operator}) results in the following update for the $i^{th}$ group of features in $v_k$ \citep{Boyd10}:
\begin{equation}
\left[v_k\right]_{{\cal{I}}_i}\leftarrow -\mu \left(1-\frac{\lambda_1}{\left\|\left[l_k\right]_{{\cal{I}}_i}\right\|_2}\right)_+ \left[l_k\right]_{{\cal{I}}_i}
\end{equation}
This can be seen as a generalization of univariate soft-thresholding to blocks of coordinates:  each group of features is either thresholded to zero or shrunk by a constant factor. The \emph{FaSTGLZ} implementation available online also handles user-specified group lasso penalties.  Moreover, \emph{FaSTGLZ} may in principle  apply to any regularizer whose proximity operator is computable in closed-form by making the appropriate adjustment in the update for $v_k$. A more extensive discussion of proximity operators and those functions $g$ that can be computed in closed-form may be found in \citet{Boyd10}.

The \emph{FaSTGLZ} algorithm may also be extended to loss functions of the form $L(\eta(w),y)$, provided $L$ is convex and twice differentiable with respect to its first argument.  In this case, the loss function for the $k^{th}$ problem would be specified by:
\begin{equation}
\sum_{i=1}^n d_k^{(i)} L(\eta^{(i)}(w_k),y_k^{(i)}) \label{eqn:gen_loss}
\end{equation}
This modification requires a change to the update for $w_k$ (see (\ref{eqn:min_Sk})) by replacing the negative log-likelihood term $\ell_k(w_k)$ in (\ref{eqn:Sk}) with the loss function above.  Approximating (\ref{eqn:gen_loss}) by a quadratic function around $\bar w_k$ results in the same structural form for the gradient and Hessian in (\ref{eqn:ell_grad}) and (\ref{eqn:ell_hess}), with $e_k^{(i)}$ and the $i^{th}$ diagonal of $R_k$, denoted $[R_k]_{ii}$, being replaced by:
\begin{eqnarray}
e_k^{(i)} &\sim& d_k^{(i)}\frac{\partial L(z,y_k^{(i)})}{\partial z}|_{z=\eta^{(i)}(\bar w_k)} \\
\left[R_k\right]_{ii} &\sim& d_k^{(i)}\frac{\partial^2 L(z,y_k^{(i)})}{\partial z^2}|_{z=\eta^{(i)}(\bar w_k)}
\end{eqnarray}
Since $L$ is assumed to be convex, the diagonal entries of $R$ are non-negative, and Newton's method will converge to the global minimum.  With these changes to $e_k$ and $R_k$, optimization then follows analogously to (\ref{eqn:min_Sk}).

\begin{comment}
\begin{algorithm}
\caption{The FaSTGLZ algorithm.  Code is also freely available at http://liinc.bme.columbia.edu/fastglz}
\begin{algorithmic}
\STATE Given: Regularization parameters $\lambda_1,\lambda_2$ and penalty parameter $\mu$.
\STATE Given: Estimate of the active set $A_k$, $k=1,\dots,K$ using the screening rules.
\STATE Initialize $\lambda_k,v_k$ as sparse vectors, nonzero only in $A_k$.
\STATE Set $l_k=\lambda_k+(1/\mu)v_k$.  Set $\beta_k=Q^Tl_k$.
\WHILE{\text{Not converged}}
\STATE Minimize $S_k(w_k,l_k)$ by solving for $\alpha_k$ in (\ref{eqn:sub_linear_system}).
\STATE Update $l_k$ and $\beta_k$:
	\begin{eqnarray}
	l_k &\leftarrow& \left(1-\frac{1}{\mu\rho}\right)l_k + \frac{2}{\mu}Q\left(\frac{1}{2\rho}\beta_k - \alpha_k\right) \\
	\beta_k &\leftarrow& \beta_k - (2/\mu)\alpha_k
	\end{eqnarray}
	where the multiplication with $Q$ is evaluated only amongst rows in the active set $A_k$.
\STATE Update $v_k$ by soft-thresholding the entries of $l_k$ belonging to $A_k$:
	\begin{equation}
	v_k = -\mu\text{ }\text{soft}(l_k,\lambda_1)
	\end{equation}
\STATE Update $l_k$ and $\beta_k$:
	\begin{eqnarray}
	l_k &\leftarrow& l_k + (2/\mu)v_k \\
	\beta_k &\leftarrow& \beta_k + (2/\mu)Q^Tv_k
	\end{eqnarray}
\ENDWHILE
\end{algorithmic}
\label{alg:alg1}
\end{algorithm}
\end{comment}

\section{Results}
\label{sec:results}

The \emph{FaSTGLZ} algorithm is most applicable to datasets in which the number of features greatly exceeds the number of examples (the $p\gg n$ problem).  This situation arises in many applications, but this section highlights examples in neuroimaging.  In this setting, the goal is to identify multi-variate patterns from a subject's brain scans that can decode various markers of cognitive state related to a task or stimulus condition \citep{Haxby01,Norman06,Sajda09}.  Before presenting the results of our algorithm, we briefly introduce the two experimental datasets:  one in functional MRI (fMRI) and one in electroencephalography (EEG).

\subsection{Data Description}
\label{sec:results_data_desc}

In the fMRI experiment, subjects participated in an auditory oddball detection task.  On each trial, the subject was presented with a standard tone (a pure $390$Hz pure tone) $80\%$ of the time, or an oddball tone (a broadband ``laser gun'' sound) the remaining $20\%$ of the time, and the subject was told to respond via button-press when an oddball stimulus was perceived.  Throughout the experiment, fMRI data were collected, and details on data preprocessing can be found in \citep{Goldman09,Walz12}.  The decoding task is to predict the stimulus category of each trial from the fMRI data.  Since there are two categories (oddball/standard), the GLZ is equivalent to logistic regression.  For each of $3$ subjects, $n=375$ trials were acquired, and features corresponded to the fMRI response from brain voxels, with $p\approx 42,000$.  

In the EEG experiment, subjects participated in a three-alternative forced choice visual discrimination task.  On each trial, the subject was presented with an image of either a face, car, or house for $200$ms, and had to respond with the category of the image by pressing one of three buttons.  To modulate the difficulty of the task, the phase coherence of the presented images were corrupted at one of two levels ($35$ or $50$), which resulted in ``easy" and ``difficult" trials.  A logistic regression GLZ could likewise be used to predict the difficulty (easy or hard) of each trial from the measured EEG data.  The feature data are spatio-temporal voltages measured across $43$ scalp electrodes and sampled at $250$Hz between stimulus onset and $600$ms post-stimulus.  Treating each electrode-timepoint pair as a feature resulted in $p=6,494$ features ($43$ electrodes $\times$ $151$ time points, plus a bias term).  The number of trials was $n=650$.  

\subsection{FaSTGLZ Results}

\begin{figure}[H]
\vskip -0.1in
\centering
\subfigure[]{
	\includegraphics[width=0.35\linewidth]{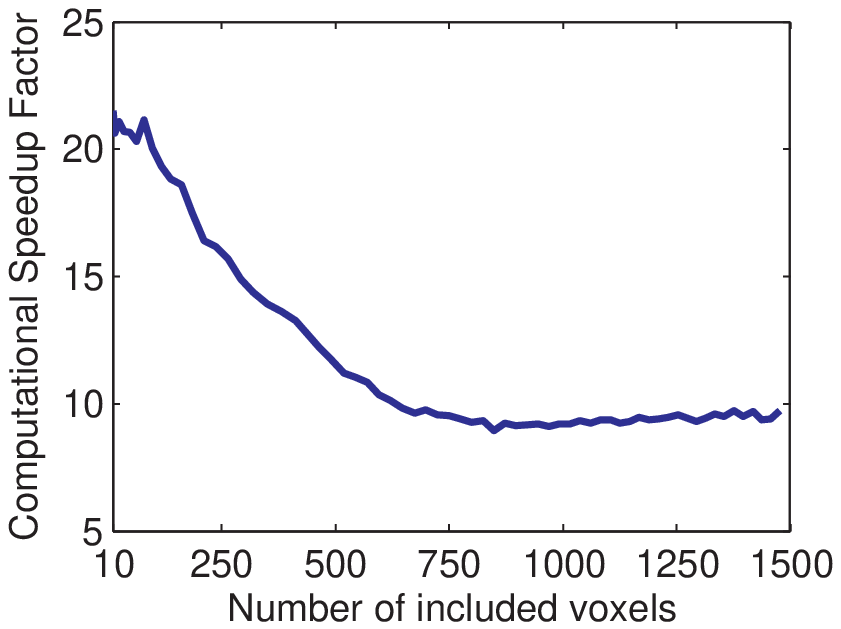}
	\label{fig:relspeed_against_1e-4}
}
\subfigure[]{
	\includegraphics[width=0.35\linewidth]{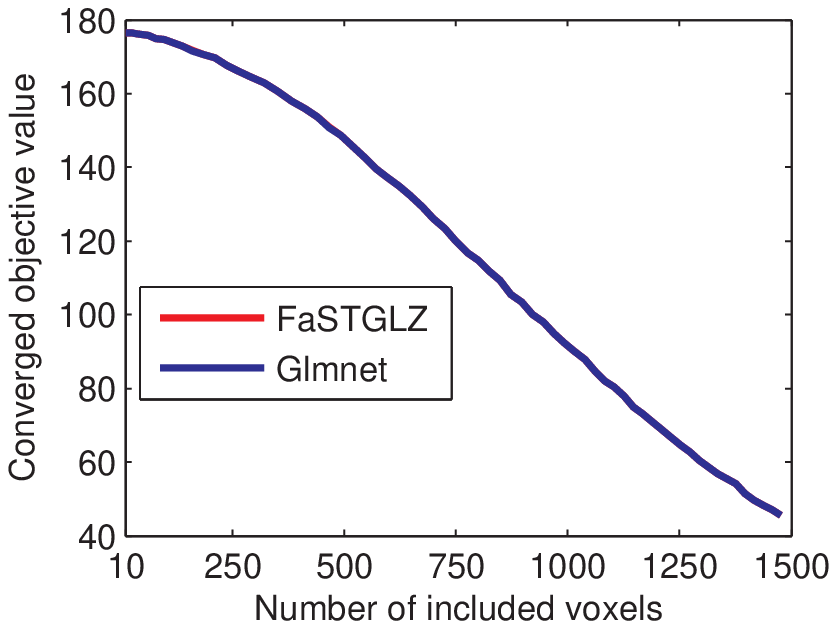}
	\label{fig:objval_against_1e-4}
}
\subfigure[]{
	\includegraphics[width=0.35\linewidth]{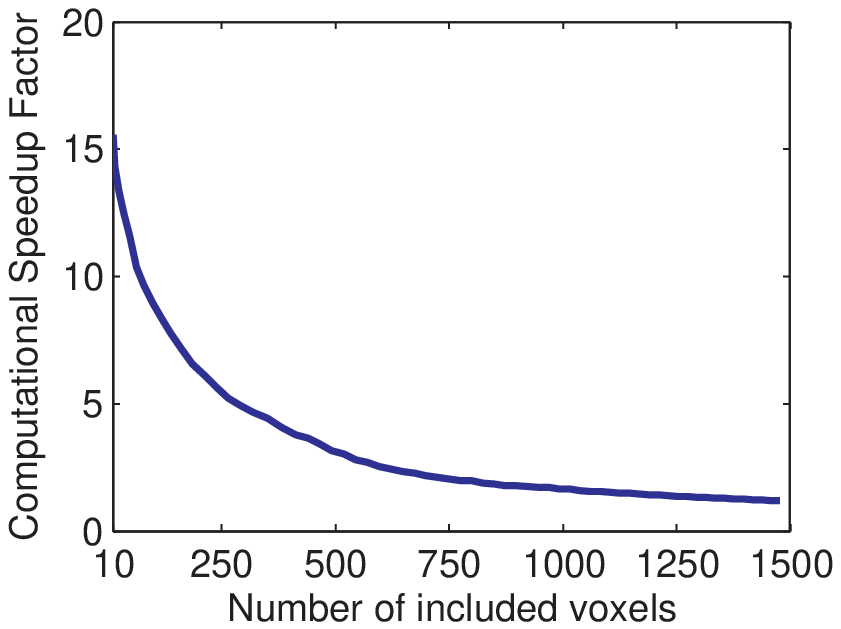}
	\label{fig:relspeed_against_1e-1}
}
\subfigure[]{
	\includegraphics[width=0.35\linewidth]{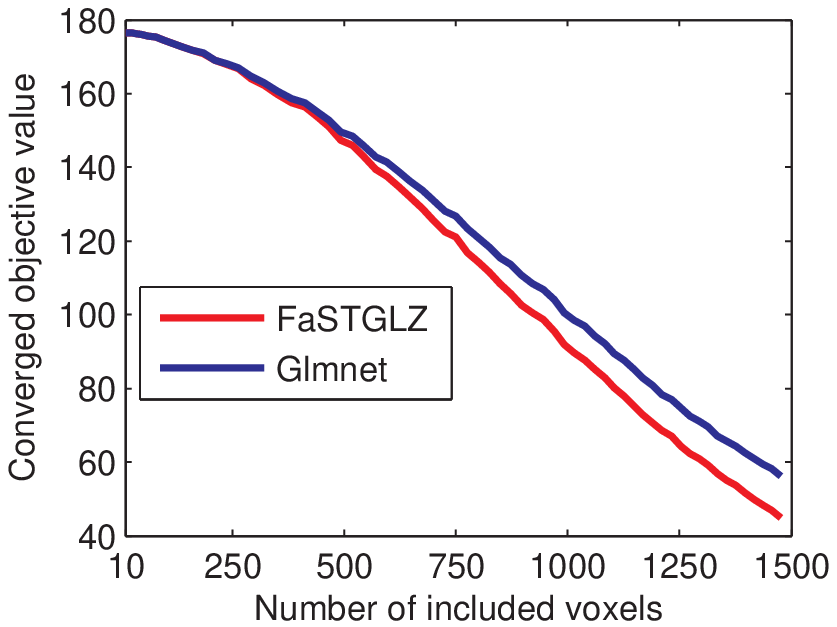}
	\label{fig:objval_against_1e-1}
}
\vskip -0.1in
\caption{Benchmarking \emph{FaSTGLZ} against \emph{glmnet} in solving a set of $K=1000$ problems on real fMRI data.  The solution to each problem was computed along a regularization path to vary the number of voxels included in the model. (a) A plot of the computational speedup factor, defined as the ratio of time required by glmnet to the time required by \emph{FaSTGLZ}, as a function of the number of voxels included in the model.  \emph{FaSTGLZ} is at least 10x faster. (b) The difference in converged objective values between \emph{FaSTGLZ} and glmnet as a function of the number of voxels included in the model.  This difference never exceeded $2\times 10^{-4}$, and the curves in (b) appear superimposed. To further evaluate the speed comparison, we tuned the convergence tolerance on glmnet so that it ran the full regularization path in roughly the same time as \emph{FaSTGLZ} (\emph{FaSTGLZ} was still 1.2x faster -- see (c)).  In this case, \emph{FaSTGLZ} produced converged objective values that were approximately $20\%$ lower than glmnet (d).}
\label{fig:relspeed}
\end{figure}

First, we benchmarked the speed of \emph{FaSTGLZ} in solving a set of $K$ problems against solving them sequentially using the popular \emph{glmnet} algorithm using coordinate descent \citep{Friedman10}.  Following \citet{Friedman10}, we parameterized the regularization parameters $(\lambda_1,\lambda_2)$ in terms of $(\alpha\lambda,0.5(1-\alpha)\lambda)$ and held $\alpha=0.7$ fixed, while $\lambda$ varied along a regularization path of $100$ values.  As an example of a significance testing problem, we trained the classifier along this regularization path for $K=1000$ permutations, and compared the time required by the two algorithms.  Figure \ref{fig:relspeed_against_1e-4} plots the computational speedup factor, defined as the ratio of time required by \emph{glmnet} to the time required by \emph{FaSTGLZ}, as a function of the average number of voxels included in the model.  \emph{FaSTGLZ} is at a minimum $10$x faster.  We verified that the relative difference in the converged objective value between the two algorithms never exceeded $2\times 10^{-4}$, and a plot of the converged objectives is shown in Figure \ref{fig:objval_against_1e-4}.  As a further check, we also increased the convergence tolerance on \emph{glmnet} so that it ran the full regularization path in roughly the same time as \emph{FaSTGLZ} (\emph{FaSTGLZ} was still $1.2$x faster -- see Figure \ref{fig:relspeed_against_1e-1}).  In this case, \emph{FaSTGLZ} produced a converged objective value that was approximately $20\%$ lower than \emph{glmnet} (see Figure \ref{fig:objval_against_1e-1}).

\begin{figure}[H]
\centering
\includegraphics[width=0.5\linewidth]{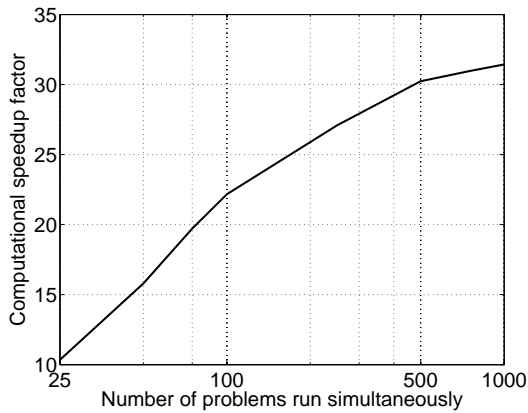}
\caption{Computational speedup factor of \emph{FaSTGLZ} with varying amounts of simultaneity relative to the non-simultaneous \emph{FaSTGLZ} applied to the same data as in Figure \ref{fig:relspeed}.}
\label{fig:relspeed2}
\end{figure}

To directly test the efficiency garnered by exploiting the shared structure across problems, we repeated the above analysis, but instead benchmarked \emph{FaSTGLZ} against itself without simultaneity, i.e., running \emph{FaSTGLZ} sequentially on each of the $K=1000$ problems.  We then ran \emph{FaSTGLZ} numerous times, varying the number of optimization problems $K_s$ that were simultaneously solved on a log scale.  Figure \ref{fig:relspeed2} plots the computational speedup factor relative to the non-simultaneous \emph{FaSTGLZ} as a function of $K_s$.  The graph has an initial rapid rise so that with $K_s=25$, the speedup is around $10$x, and there is an inflection point at around $K_s=500$ simultaneous problems (speedup of $30$x), after which point there is a diminishing rate of return in efficiency.  Interestingly, the base \emph{FaSTGLZ} algorithm without simultaneity is about $3$x slower than glmnet.  This further emphasizes that the empirical computational improvement relative to glmnet in the previous example is directly a result of exploiting the shared structure across problems.

\begin{figure}[H]
\centering
\includegraphics[width=0.5\linewidth]{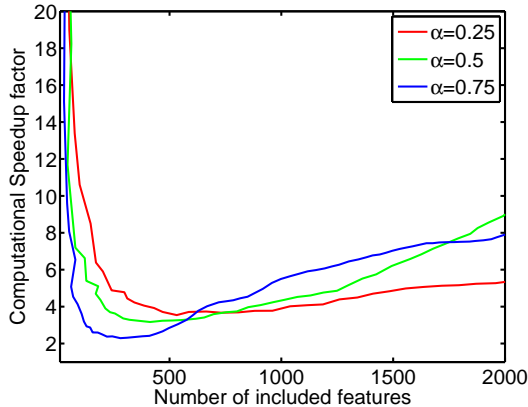}
\caption{Benchmarking \emph{FaSTGLZ} against \emph{glmnet} in solving a set of $K=1000$ bootstrapping problems on real EEG data.  The solution to each problem was computed along a regularization path to vary the number of features included in the model.}
\label{fig:eeg_speedup}
\end{figure}

We also benchmarked \emph{FaSTGLZ} on the EEG example dataset described in Section \ref{sec:results_data_desc}.  In this instance, we compared the speed of \emph{FaSTGLZ} in solving a set of $K=1000$ problems derived via bootstrapping against solving them sequentially using \emph{glmnet}.  For each problem, the training set was derived by sampling with replacement from the set of $n=650$ trials.  The weight $d_k^{(i)}$ assigned to trial $i$ for problem $k$ was then set to the number of times that trial was sampled in the bootstrap.  The regularization parameter $\lambda$ was varied along a regularization path of $150$ values for each of $3$ values of $\alpha\in\{0.25,0.5,0.75\}$.  Figure \ref{fig:eeg_speedup} plots the computational speedup factor of \emph{FaSTGLZ} relative to \emph{glmnet} as a function of the number of features included in the model for each value of $\alpha$.  Again, the relative difference in converged objective values was observed to never exceed $10^{-4}$.
 
\section{Conclusion}
\label{sec:conclusion}

We presented the fast simultaneous training of generalized linear models (\emph{FaSTGLZ}) algorithm and demonstrated its significant speedup in computational efficiency when analyzing high-dimensional real-world datasets. \emph{FaSTGLZ} enables efficiently training families of elastic-net regularized GLZ models that may arise from bootstrapping, cross-validation, and permutation testing.  We also provided a discussion on extensions of the algorithm to other regularizers and loss functions, including group lasso.  Moreover, by appropriately setting the elastic net regularization parameters, ridge and lasso are also accommodated as special cases.

%We then provided an example application of our algorithm on a real auditory and visual oddball fMRI dataset.  The computational speed-up of our algorithm allowed us to run intensive permutation tests to evaluate statistical significance of both the prediction accuracy and brain map reproducibility of discriminative classifiers derived from the fMRI data.  We also suggested a model selection strategy for tuning regularization parameters that takes into account a combination of the prediction accuracy and the brain map reproducibility of the derived classifier.  This was shown to greatly improve interpretability of the models over the standard approach of maximizing prediction accuracy alone.

The \emph{FaSTGLZ} algorithm is particularly efficient when the number of examples $n$ is limited, but the number of features $p$ may be very large (the $p\gg n$ problem).  This is often the case in many real-world scenarios, including neuroimaging datasets such as EEG, MRI, and DTI, as well as genetic datasets.

From the algorithmic perspective, there are a number of future research directions.  \citet{Boyd10} showed that ADMM can be coupled with distributed optimization to effectively handle very large-scale datasets (large $p$ and $n$).  Connecting this approach with our \emph{FaSTGLZ} algorithm could potentially be very fruitful.  Another potential research direction is in making simultaneous versions of other machine learning algorithms. For example, faster variants of ADMM have been presented recently in \citet{Goldfarb09,Goldstein12}. Adapting these algorithms to simultaneous versions could further improve computational efficiency. 

\subsubsection*{Acknowledgments}

This work was supported by National Institutes of Health grant R01-MH085092, the National Science Foundation Graduate Research Fellowship Program, and by the Army Research Laboratories under Cooperative Agreement Number W911NF-10-2-0022.  The views and conclusions are those of the authors and should not be interpreted as representing the official policies, either expressed or implied, of the Army Research Laboratory or the U.S. Government. 

We thank Jordan Muraskin, Robin Goldman, and Eftychios Pnevmatikakis for their fruitful discussions and suggestions.  We also thank Glenn Castillo and Stephen Dashnaw for their assistance with EEG-fMRI data acquisition.

\section*{Supplementary Material}
\subsection*{Proof of convergence of simultaneous Newton's method iterative solver (see \S \ref{sec:simult_newt_meth})}

Here we show that by taking the template matrix to be $M=ZRZ^T+2\lambda_2 I$, with $R=\max(R_1,\dots,R_K)$ the element-wise maximum, then the iterative method:
\begin{equation}
\alpha_k^{(t+1)} = M^{-1}Z(R-R_k)Z^T\alpha_k^{(t)} + M^{-1}Zb_k \label{eqn:iter_meth}
\end{equation}
converges to the true solution $G_k^{-1}Zb_k$ of (\ref{eqn:linear_sys_reduced}) for each $k=1,\dots,K$.

First note that if (\ref{eqn:iter_meth}) does converge for some $t\geq t^*$, then $\alpha_k^{(t^*)}$ is the solution to (\ref{eqn:linear_sys_reduced}).  To see this, note that upon convergence $\alpha_k^{(t^*+1)}=\alpha_k^{(t^*)}$, and plugging that into (\ref{eqn:iter_meth}), we have:
\begin{eqnarray}
\alpha_k^{(t^*)} &=& M^{-1}Z(R-R_k)Z^T\alpha_k^{(t^*)} + M^{-1}Zb_k \\
			&=& M^{-1}(M-G_k)\alpha_k^{(t^*)} + M^{-1}Zb_k \\
			&=& \alpha_k^{(t^*)} - M^{-1}G_k\alpha_k^{(t^*)} + M^{-1}Zb_k \\
			&=& G_k^{-1}Zb_k
\end{eqnarray}
where the second step above used the fact that $Z(R-R_k)Z^T=(M-G_k)$.

To prove convergence, we must show that the spectral radius of $M^{-1}Z(R-R_k)Z^T$ is less than one \cite{Young71}. To simplify notation, let $N_k=M-G_k=Z(R-R_k)Z^T$.  Since the eigenvalues of $M^{-1}N_k$ correspond to the generalized eigenvalues $\lambda M\alpha=N_k\alpha$, we may compute the spectral radius $\rho(M^{-1}N_k)$ by maximizing the generalized Rayleigh quotient:
\begin{eqnarray}
\rho(M^{-1}N_k) &=& \max_\alpha \left| \frac{\alpha^TN_k\alpha}{\alpha^TM\alpha} \right| \label{eqn:gen_rayleigh}
\end{eqnarray}
Taking $R=\max(R_1,\dots,R_K)$ assures that $N_k$ and $M$ are positive semi-definite matrices for all $k$.  As a result, the numerator and denominator of (\ref{eqn:gen_rayleigh}) are real, non-negative numbers.  As a result, constraining $\rho(M^{-1}N_k)<1$ requires that for all vectors $\alpha$:
\begin{eqnarray}
\alpha^TN_k\alpha &<& \alpha^TM\alpha \\
\alpha^TZ(R-R_k)Z^T\alpha &<& \alpha^TZRZ^T\alpha + 2\lambda_2 \\
\alpha^TZR_kZ^T\alpha + 2\lambda_2 &>& 0 \label{eqn:converg_condition}
\end{eqnarray}
Thus, (\ref{eqn:iter_meth}) converges provided that $\alpha^TZR_kZ^T\alpha + 2\lambda_2>0$ for all $\alpha$, which is guaranteed since $ZR_kZ^T$ is PSD and $\lambda_2>0$.

\subsection*{Equivalence of ADMM Procedures ((\ref{eqn:admm_step1})-(\ref{eqn:admm_step4})) and ((\ref{eqn:min_Sk})-(\ref{eqn:admmmod_step4})) (see \S \ref{sec:fastglz_with_sparsity})}

Here we show that the standard ADMM procedure (\ref{eqn:admm_step1})-(\ref{eqn:admm_step4}) is equivalent to the modified procedure (\ref{eqn:min_Sk})-(\ref{eqn:admmmod_step4}).  First, we expand (\ref{eqn:admm_step1}) as:
\begin{eqnarray*}
\arg\min_{w_k}{\cal{L}}_k(w_k,v_k) &=& \arg\min_{w_k} \ell_k(w_k) + \lambda_2||w_k||_2^2 - \lambda_k^Tw_k + \frac{1}{2\mu}||w_k-v_k||_2^2 \\
	&=& \arg\min_{w_k} \ell_k(w_k) + (\lambda_2+\frac{1}{2\mu})||w_k||_2^2 - (\lambda_k+\frac{1}{\mu}v_k)^Tw_k \\
	&=& \arg\min_{w_k} S_k(w_k,l_k)
\end{eqnarray*}
where $l_k$ is given by $l_k=\lambda_k+(1/\mu)v_k$.  Thus, (\ref{eqn:admm_step1}) is equivalent to (\ref{eqn:min_Sk}).

Upon updating $\lambda_k$ in (\ref{eqn:admm_step2}) and comparing with the update for $l_k$ in (\ref{eqn:admmmod_step2}), we have that they are then related by $l_k=\lambda_k-(1/\mu)w_k$.  Now, we expand the update for $v_k$ in (\ref{eqn:admm_step3}) as:
\begin{eqnarray}
\arg\min_{v_k}{\cal{L}}_k(w_k,v_k) &=& \arg\min_{v_k} \lambda_1||v_k||_1 + \lambda_k^Tv_k + \frac{1}{2\mu}||w_k-v_k||_2^2 \\
	&=& \arg\min_{v_k} \lambda_1||v_k||_1 + (\lambda_k - \frac{1}{\mu}w_k)^Tv_k + \frac{1}{2\mu} ||v_k||_2^2 \\
	&=& \arg\min_{v_k} \mu\lambda_1||v_k||_1 + \mu l_k^Tv_k + \frac{1}{2}||v_k|_2|^2 \\
	&=& \arg\min_{v_k} \mu\lambda_1 ||v_k||_1 + \frac{1}{2}||v_k + \mu l_k||_2^2 \\
	&=& -\text{soft}(\mu l_k,\mu\lambda_1\mathbf{1}) \\
	&=& -\mu \text{ soft}(l_k,\lambda_1\mathbf{1})
\end{eqnarray}
Thus, (\ref{eqn:admm_step3}) is equivalent to (\ref{eqn:admmmod_step3}).

Finally, after $\lambda_k$ is updated in (\ref{eqn:admm_step4}) and $l_k$ is updated in (\ref{eqn:admmmod_step4}), we again have that $l_k=\lambda_k+(1/\mu)v_k$, and the iterations repeat.

\subsection*{Proof of (\ref{eqn:wk}) (see \S \ref{sec:fastglz_with_sparsity})}

Here we prove that the solution $w_k$ to the linear system in (\ref{eqn:Sk}) is given by (\ref{eqn:wk}).  

First, we establish that the solution $w_k$ must lie in $\text{span}(Q)\cup \text{span}(l_k)$.  To show this, take any vector $a\perp \text{span}(Q)\cup \text{span}(l_k)$.  Then taking the inner product of $a$ with both sides of (\ref{eqn:Sk}), we obtain:
\begin{eqnarray}
a^T(H_k+2\rho I)w_k &=& a^T(H_k\bar w_k - \nabla\ell_k + l_k) \\
(2\rho) a^Tw_k &=& 0
\end{eqnarray}
The above follows from the fact that $\nabla\ell_k\in\text{span}(Q)$ and for any $b$, $H_kb\in\text{span}(Q)$ (see equations (\ref{eqn:ell_grad}) and (\ref{eqn:ell_hess})).  Therefore, $a\perp w_k$, implying that $w_k\in\text{span}(Q)\cup\text{span}(l_k)$.

At this point, we impose a change of basis by expressing $w_k=Q\alpha_k+Q^{\perp}\gamma_k$, where the columns of $Q^{\perp}$ span the orthogonal complement of $\text{span}(Q)$.  This gives us:
\begin{eqnarray}
\left[\begin{array}{c}Q^T\\{Q^{\perp}}^T\end{array}\right](H_k+2\rho I)\left[\begin{array}{cc} Q & Q^{\perp}\end{array}\right]\left[\begin{array}{c}\alpha_k\\\gamma_k\end{array}\right] = \left[\begin{array}{c}Q^T\\{Q^{\perp}}^T\end{array}\right](H_k\bar w_k - \nabla\ell_k + l_k) \\
\left[\begin{array}{cc} ZR_kZ^T + 2\rho I & 0 \\ 0 & 2\rho I\end{array}\right]\left[\begin{array}{c}\alpha_k \\ \gamma_k\end{array}\right] = \left[\begin{array}{c} Zb_k + Q^T l_k \\ {Q^{\perp}}^T l_k\end{array}\right]
\end{eqnarray}
Thus, $\gamma_k=\frac{1}{2\rho}{Q^{\perp}}^T l_k$ and $\alpha_k$ is the solution to:
\begin{eqnarray}
(ZR_kZ^T + 2\rho I)\alpha_k = Zb_k + Q^T l_k
\end{eqnarray}
which is the same as that given in (\ref{eqn:sub_linear_system}).  Substituting into our original expansion for $w_k$: $w_k=Q\alpha_k+Q^{\perp}\gamma_k$, we obtain:
\begin{eqnarray}
w_k &=& Q\alpha_k + Q^{\perp}{Q^{\perp}}^T\gamma_k \\
	&=& Q\alpha_k + \frac{1}{2\rho}(I - P_{Q})l_k
\end{eqnarray}
which completes the proof.

\bibliography{references}

\end{document}